\documentclass[10pt,twocolumn,letterpaper]{article}

\usepackage[pagenumbers]{cvpr} 
\usepackage{multirow}
\usepackage{amsmath}
\usepackage{amssymb}
\usepackage{booktabs}
\usepackage[accsupp]{axessibility} 

\usepackage[dvipsnames]{xcolor}

\definecolor{cvprblue}{rgb}{0.21,0.49,0.74}
\usepackage[pagebackref,breaklinks,colorlinks,citecolor=cvprblue]{hyperref}

\def\paperID{} 
\def\confName{CVPR}
\def\confYear{2024}

\title{Bridging the Gap Between End-to-End and Two-Step Text Spotting}

\author{
Mingxin Huang\textsuperscript{1}
\quad Hongliang Li\textsuperscript{1}
\quad Yuliang Liu\textsuperscript{2}
\quad Xiang Bai\textsuperscript{2}
\quad Lianwen Jin\textsuperscript{1,3}$^*$
\\
\textsuperscript{1}{South China University of Technology} 
\quad \textsuperscript{2}{Huazhong University of Science and Technology}
\\
\textsuperscript{3}{INTSIG-SCUT Joint Lab on Document Analysis and Recognition}
\\
{\tt\small eelwjin@scut.edu.cn}
}

\begin{document}
\maketitle
\begin{abstract}
Modularity plays a crucial role in the development and maintenance of complex systems. While end-to-end text spotting efficiently mitigates the issues of error accumulation and sub-optimal performance seen in traditional two-step methodologies, the two-step methods continue to be favored in many competitions and practical settings due to their superior modularity. In this paper, we introduce Bridging Text Spotting, a novel approach that resolves the error accumulation and suboptimal performance issues in two-step methods while retaining modularity. To achieve this, we adopt a well-trained detector and recognizer that are developed and trained independently and then lock their parameters to preserve their already acquired capabilities. Subsequently, we introduce a Bridge that connects the locked detector and recognizer through a zero-initialized neural network. This zero-initialized neural network, initialized with weights set to zeros, ensures seamless integration of the large receptive field features in detection into the locked recognizer. Furthermore, since the fixed detector and recognizer cannot naturally acquire end-to-end optimization features, we adopt the Adapter to facilitate their efficient learning of these features. We demonstrate the effectiveness of the proposed method through extensive experiments: Connecting the latest detector and recognizer through Bridging Text Spotting, we achieved an accuracy of 83.3\% on Total-Text, 69.8\% on CTW1500, and 89.5\% on ICDAR 2015. The code is available at \url{https://github.com/mxin262/Bridging-Text-Spotting}.

\end{abstract}
\let\thefootnote\relax\footnotetext{$^*$Corresponding author.}

\section{Introduction}
\label{sec:intro}

Text spotting, as a critical technology for reading text in natural scenes, has garnered significant attention in recent years, owing to its diverse real-world applications, including autonomous driving~\cite{9551780}, intelligent navigation~\cite{rong2016guided,wang2015bridging}, and visual information extraction~\cite{cao2023attention,van2023document,kuang2023visual}. 
Traditional two-step text spotting separates text detection and recognition into two independent models~\cite{liao2017textboxes}. 
In the first step, the focus is on detecting the text regions within a natural scene image. Once the text regions have been located, the second step involves cropping these regions within the image and employing a recognition model to recognize the text contained within these cropped regions. 

\begin{figure}[t!]
    \centering
    \includegraphics[width=\linewidth]{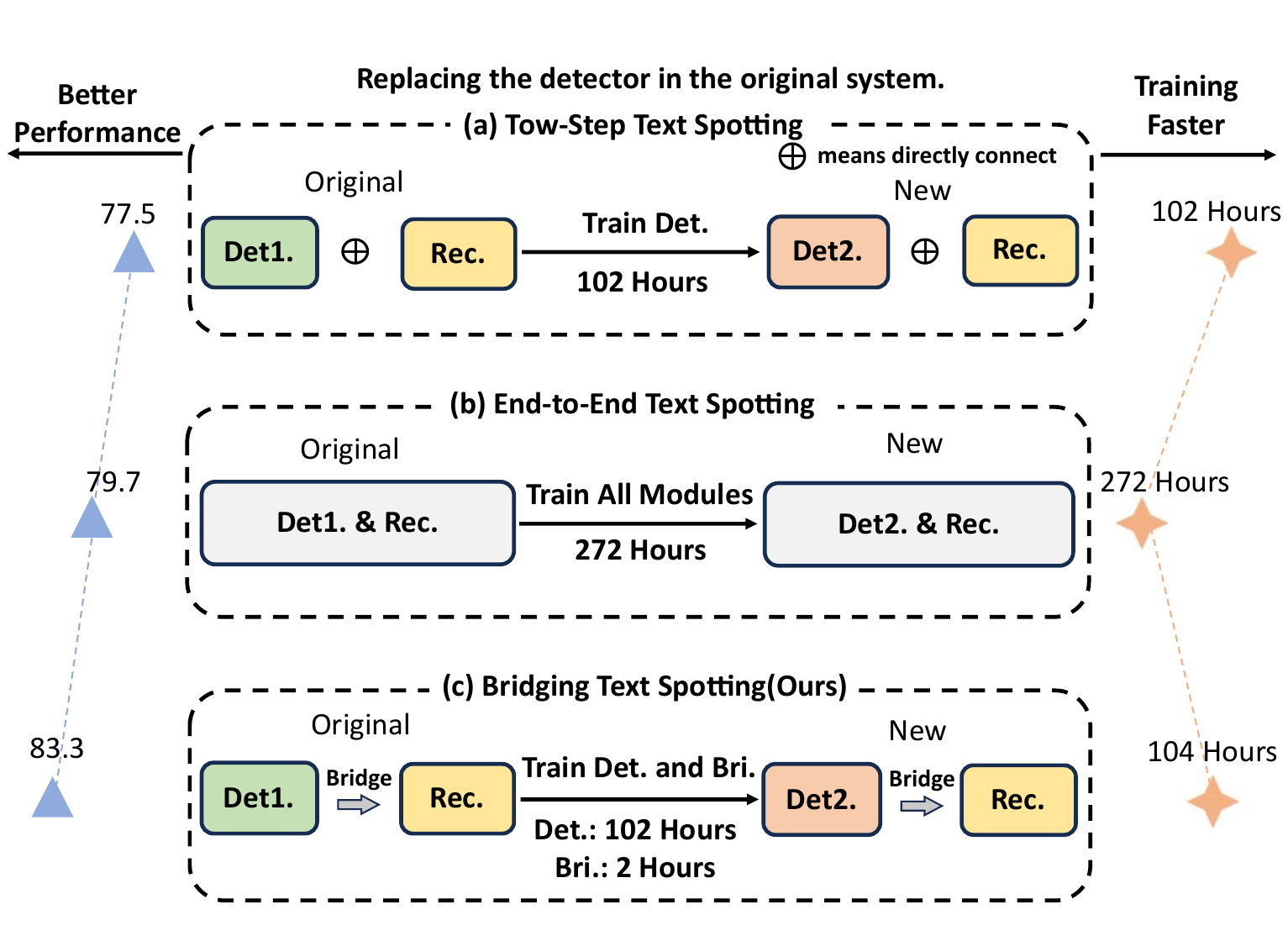}
    \caption{Comparison between the proposed paradigm with existing text spotting paradigms. Our pipeline achieves better performance with high modularity. We adopt the latest detector~\cite{zhang2023arbitrary} and text spotter~\cite{ye2023deepsolo} to test the training time of the two-step and end-to-end methods, respectively. The training time is evaluated on the RTX-3090. Det1. means the original detector. Det2. means the new detector. Rec. means the text recognizer. Bri. mean the proposed Bridge.
    }
    \label{fig:intro}
\end{figure}

Recently, many researchers have integrated text detection and recognition within an end-to-end trainable framework, aiming to address the issues of sub-optimal performance and error propagation~\cite{li2017towards,liao2019mask,liu2021abcnetv2,huang2022swintextspotter}. Dominant end-to-end text spotting methods mainly follow the
detection-by-recognition paradigm~\cite{li2017towards,liu2018fots,liao2019mask,liao2020mask,liu2020abcnet}. In these approaches, text detectors are initially used to locate text instances, followed by a Region-of-Interest (RoI) operation that extracts features from the shared backbone for recognition. These methods often require the incorporation of numerous heuristic designs involving RoI operations and post-processing steps~\cite{zhang2022text,ye2023deepsolo}. Inspired by the Transformer~\cite{vaswani2017attention}, recent advances~\cite{zhang2022text,kittenplon2022towards,ye2023deepsolo,kil2023towards} develop a Transformer-based text spotting framework to avoid the RoI operation and post-processing steps. Besides, some researchers~\cite{huang2022swintextspotter,ye2023deepsolo,huang2023estextspotter} attempt to enhance the synergy between the detection and recognition. 

Despite the significant progress in end-to-end text spotting, many competitions~\cite{zhang2019icdar,sun2019icdar,wu2023icdar,long2023icdar} and practical applications~\cite{du2020pp,du2021pp,li2022pp} still favor the two-step text spotting. One crucial reason for this preference is the high modularity inherent in two-step methods. Modularity refers to breaking up a complex system into discrete pieces. Highly modular text spotting systems allow simultaneous development and independent maintenance of detectors and recognizers, with the flexibility to adjust the amount of training data for each module based on specific requirements~\cite{baldwin2000design,langlois2002modularity}.
To more intuitively illustrate the advantages of modularity, consider a scenario where we need to replace the detector in the original system. As depicted in Fig.~\ref{fig:intro}, the two-step text spotting approach simply requires training a new detector, which can be completed in approximately 102 hours. In contrast, the end-to-end text spotting method necessitates retraining the entire system, which consumes roughly 272 Hours. Furthermore, when aiming to improve the performance of the detection by increasing training data, unlike two-step methods that only necessitate labeling for detection annotations, end-to-end methods require labeling for both detection and recognition annotations, resulting in higher resource consumption than two-step methods.

In this paper, we introduce a new paradigm for text spotting, termed Bridging Text Spotting, which addresses the issues of error accumulation and sub-optimal performance in two-step methods while retaining modularity.
Specifically, Bridging Text Spotting adopts a well-trained detector and recognizer and then locks them to maintain their already acquired capabilities. 
Then, we propose a \textit{Bridge} to integrate the locked detector and recognizer into a trainable framework by incorporating the large receptive field features from the detection into the locked recognizer. 
To prevent the recognizer from misinterpreting the detection feature as noise in the early stages of training, we initialize the weights of the input and output layer in the \textit{Bridge} to zeros. 
Additionally, as the locked detector and recognizer do not inherently possess end-to-end optimization features, we adopt the Adapter~\cite{houlsby2019parameter} to facilitate their efficient learning of these features. When transitioning to a new scenario, the Bridging Text Spotting simply requires training a new detector and \textit{Bridge} with the Adapter. 
It's worth noting that training the \textit{Bridge} with the Adapter is efficient, as demonstrated in the bottom part of Fig.~\ref{fig:intro}. Benefiting from the utilization of the well-trained detector and recognizer, \textit{Bridge} with Adapter eliminates the need for extensive data and enables a rapid completion of the training process. 

In conclusion, the main contributions are three-fold:

\begin{itemize}
\item 
We introduce a new paradigm for text spotting, termed Bridging Text Spotting, which addresses the issues of error accumulation and sub-optimal performance in two-step spotting while retaining modularity.

\item 
We propose a \textit{Bridge} with the Adapter that enables the well-trained detector and recognizer to learn the end-to-end optimization features based on their already acquired capabilities.

\item 
Experiments demonstrate the effectiveness of the proposed Bridging Text Spotting: 1) Connecting the latest detector and recognizer through Bridging Text Spotting, we achieved an accuracy of 83.3\% on Total-Text, 69.8\% on CTW1500, and 89.5\% on ICDAR 2015; 2) Bridging Text Spotting can achieve an average improvement of 4.4\% across multiple combinations of detectors and recognizers.

\end{itemize}

\begin{figure*}[h]
    \centering
    \includegraphics[width=1.0\linewidth]{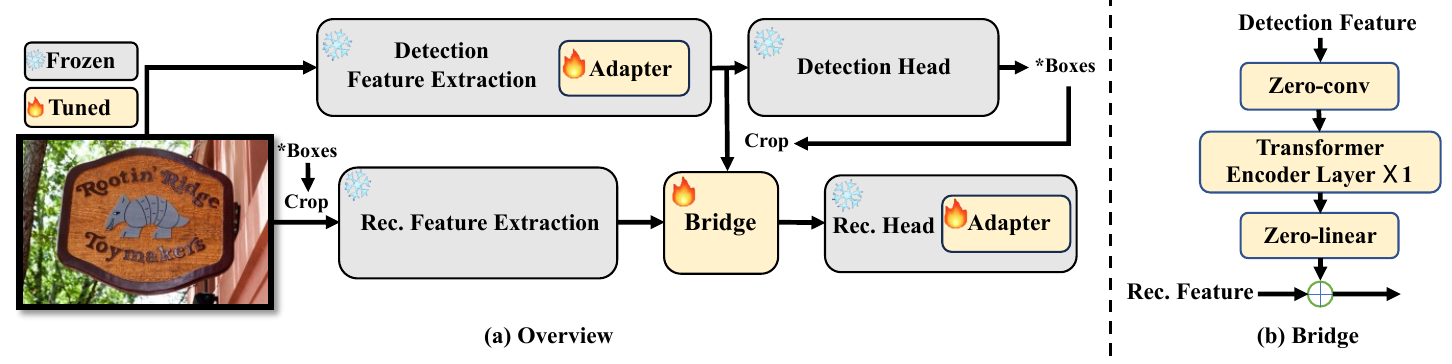}
    \caption{
    The overall architecture of bridging text spotting. Rec. means the recognition. Crop represents the crop operation. The predictions of the detector are used to crop the text regions.
    }
    \label{fig:frame}
\end{figure*}

\section{Related Work}
\label{sec:rela}

Over the past few decades, the advent of deep learning techniques has significantly advanced the field of scene text spotting. Text spotting methods can be broadly classified into two main categories: two-step text spotting and end-to-end text spotting.

\textbf{Two-Step Text Spotting}. 
Two-step text spotting involves performing detection and recognition through two separate models. The detection model initially locates the text regions, and then the recognition model recognizes the text within these regions. In recent years, Wang \etal \cite{wang2011end} detect characters by a sliding-window-based detector and subsequently classify each character. Jaderberg \etal \cite{jaderberg2016reading} introduce a method that first detects text instances by generating holistic text proposals with high recall and then recognizes the text content using a word classifier. Liao \etal propose TextBoxes++~\cite{liao2018textboxes++}, which incorporate a single-shot detector~\cite{liao2017textboxes} and a text recognizer~\cite{shi2016end} in a two-step process. Two-step text spotting methods have a high modularity that allows independent development and maintenance of detectors and recognizers. However, the advancement of two-step text spotting faces constraints due to the accumulation of errors and sub-optimal performance issues~\cite{li2017towards,liao2019mask}.

\textbf{End-to-end Text Spotting}. 
To solve the error accumulation and sub-optimal performance issues, researchers have recently attempted to integrate detection and recognition within an end-to-end trainable framework. Li~\etal~\cite{li2017towards} integrate detection and recognition into a unified scene text spotting framework, primarily focusing on horizontal text. In order to handle oriented text, various sampling techniques, including RoIRotate~\cite{liu2018fots} and Text-Align~\cite{he2018end}, have been developed to convert the oriented text into a horizontal grid. The Mask TextSpotter series~\cite{lyu2018mask,liao2019mask,liao2020mask} further utilize character segmentation to handle the arbitrarily-shaped text. Concurrently, arbitrarily-shaped sampling techniques such as RoISlide~\cite{feng2019textdragon} and BezierAlign~\cite{liu2020abcnet,liu2021abcnetv2} are built to rectify the curve texts. Similarly, Wang \etal~\cite{wang2020all} rectify curve texts by the Thin-Plate-Spline~\cite{bookstein1989principal}. In comparison, ~\cite{qin2019towards,wang2021pan++} use the RoI Masking to connect the detector and recognizer. In order to enhance text recognition performance, Fang \etal~\cite{fangabinet++} introduce a language model~\cite{fang2021read} for text spotting. Additionally, GLASS~\cite{ronen2022glass} introduces a global-to-local attention module aimed at improving the capability to read text under varying scales. SRSTS~\cite{wu2022decoupling} reduces the dependence of recognition on detection through a self-reliant sampling recognition branch.

With the exceptional performance exhibited by the Transformer~\cite{vaswani2017attention}, researchers have started to explore its application in the field of text spotting. Zhang \etal \cite{zhang2022text} adopt a dual decoder framework to represent detection and recognition, respectively. TTS~\cite{kittenplon2022towards} add an RNN-based recognition head on Deformable DETR~\cite{zhu2020deformable}. SwinTextSpotter~\cite{huang2022swintextspotter,huang2024swintextspotter} propose a Recognition Conversion to enable the back-propagation of recognition information to the detector. To enhance the coherence of detection and recognition, Ye \etal~\cite{ye2023deepsolo} develop shared point queries for detection and recognition within a single decoder. SPTS~\cite{peng2022spts,liu2023spts} and UNITS~\cite{kil2023towards} treat text spotting as a sequence generation problem. ESTextSpotter~\cite{huang2023estextspotter} further proposes a framework to achieve explicit synergy between two tasks.

While end-to-end text spotting effectively addresses the issues of error accumulation and sub-optimal performance in traditional two-step methodologies, it faces a limitation in leveraging a substantial amount of data that solely comprises detection or recognition annotations. Additionally, end-to-end text spotting cannot directly utilize well-trained detectors and recognizers. Therefore, in many competitions and practical applications, two-step text spotting continues to be the preferred choice due to the high modularity~\cite{du2021pp,li2022pp,zhang2019icdar,sun2019icdar,wu2023icdar}.

\section{
Methodology
}
Bridging text spotting represents a fresh methodology in the realm of text spotting. It provides a novel solution to address the issues of error propagation and sub-optimal performance in two-step text spotting while preserving modularity. Within the framework of Bridging Text Spotting, the detector and recognizer can be independently developed and trained. Subsequently, they are unified via the proposed \textit{Bridge}.

\subsection{Overall Architecture}
\label{sec:overall}
The overall architecture is depicted in Fig.~\ref{fig:frame}.
Initially, we employ a well-trained detector and recognizer, both developed and trained independently. This independent development and training provide flexibility in adjusting training data and structures for individual modules as needed. Subsequently, the parameters of both the detector and recognizer are locked to preserve their already acquired capabilities. Given a scene text image, we send it to the trained detector to locate text instances. Subsequently, the detection results are employed to crop the corresponding regions within the features from the detection backbone and the image. We directly use rectangles to crop the regions. These cropped regions in the image, denoted as $\mathbf{C_i}$, are then fed into the recognition backbone to extract the features. Then, the output from the recognition backbone is sent into the Bridge along with the cropped features $\mathbf{C_f}$ from the detection backbone. In the final step, the output from the Bridge is forwarded to the recognition head, which generates the final recognition results. Additionally, Adapter~\cite{houlsby2019parameter} is integrated into the detection feature extraction and recognition head, facilitating the learning of end-to-end optimization features in both the frozen detector and recognizer.

\begin{figure}[t]
    \centering
    \includegraphics[width=0.8\linewidth]{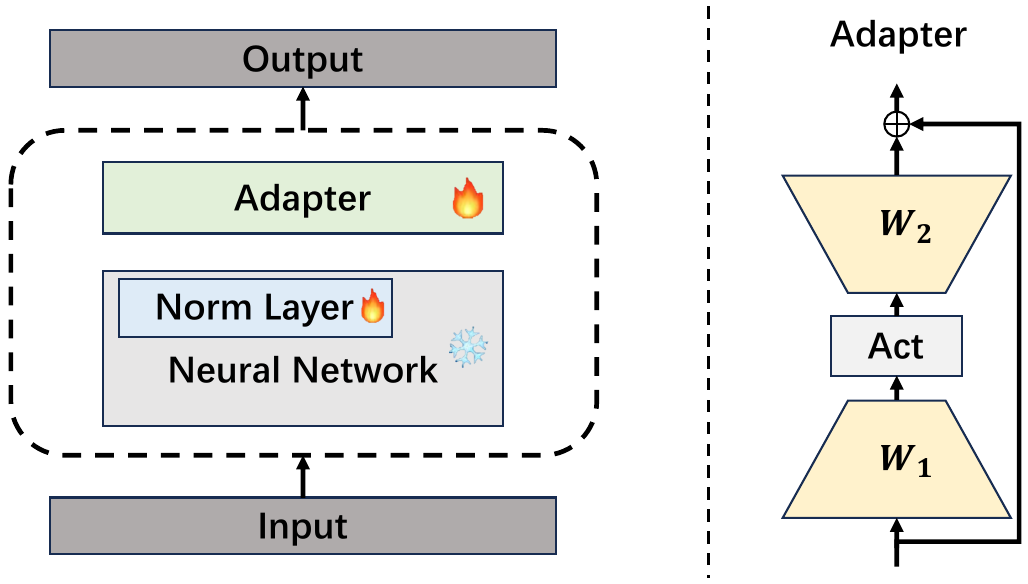}
    \caption{
    Illustration of Adapter. The neural network refers to the fundamental building blocks of a neural network, such as a multi-head attention block or a transformer block. $\mathbf{W_1}$ and $\mathbf{W_2}$ represent the linear layer. Act means the activation function. All normalization layers in the recognizer are used to tune.
    }
    \label{fig:adapter}
\end{figure}

\subsection{Bridge}
We propose a Bridge to connect the well-trained detector and recognizer, addressing the challenges of error accumulation and sub-optimal performance in two-step spotting while preserving modularity. Note that, we lock (freeze) the parameters of the well-trained detector and the recognizer to preserve their already acquired capabilities. Suppose $R(·; \theta_{rb})$ is the recognizer's backbone with parameters $\theta_{rb}$ and $R(·; \theta_{rh})$ is the recognizer's head with parameters $\theta_{rh}$. Similarly, suppose $D(·; \theta_{db})$ is the detector's backbone with parameters $\theta_{db}$ and $D(·; \theta_{dh})$ is the detector's head with parameters $\theta_{dh}$. Given the input image $\mathbf{I}$, the detection process is as follows:
\begin{equation}\small
\mathbf{F_{det}} = D(\mathbf{I}; \theta_{db}) \,,
\end{equation}
\begin{equation}\small
\mathbf{P_{det}} = D(\mathbf{F_{det}}; \theta_{dh}) \,,
\end{equation}
where $\mathbf{F_{det}}$ represents the features generated by the detection backbone. $\mathbf{P_{det}}$ represents the predictions of the detector. After obtaining the predictions from the detector, we proceed to extract the corresponding regions from the images and features generated by the detection backbone:
\begin{equation}\small
\mathbf{C_f} = Crop(\mathbf{F_{det}}, \mathbf{P_{det}}) \,,
\end{equation} 
\begin{equation}\small
\mathbf{C_i} = Crop(I, \mathbf{P_{det}}) \,,
\end{equation} 
\begin{equation}\small
\mathbf{F_i} = R(\mathbf{C_i}; \theta_{rb}) \,,
\end{equation}
where $Crop$ represents the crop operation. $\mathbf{F_i}$ represents the recognition features generated by the recognition backbone. The recognition features $\mathbf{F_i}$ and the cropped features from detection backbone $\mathbf{C_f}$ are sent to the Bridge. Inspired by ~\cite{zhang2023adding}, we design a zero-initialized convolution and zero-initialized linear layers, whose weight and bias are both initialized to zeros, denoted $Z_c(·; ·)$ and $Z_l(·; ·)$. The process in the Bridge can be formulated as follows:
\begin{equation}\small
\mathbf{F_r} = \mathbf{F_i} + Z_l(Tr(Z_c((\mathbf{C_f}+PE):\theta_{zc})):\theta_{zl}) \label{equation:zl} \,,
\end{equation}
where $Tr$ represents the Transformer encoder. $\theta_{zc}$ and $\theta_{zl}$ denote the parameters of the zero-initialized convolution and linear layers, which include the weight $\mathbf{W}$ and bias $\mathbf{B}$. $\mathbf{F_r}$ denotes the output of Bridge. During the initial training stage, the weight and bias parameters of the zero-initialized convolution and linear layers are initialized to zero. Consequently, this causes the component $Z_c(·; ·)$ and $Z_l(·; ·)$ in  eq.~\ref{equation:zl} to yield a result of zero. Then, eq.~\ref{equation:zl} is changed as follows:
\begin{equation}\small
\mathbf{F_r} = \mathbf{F_i} \,.
\end{equation}
In this way, the recognition head will not be disturbed by the sudden addition of features in the initial training phase. Although the weight and bias parameters of the convolution and linear layer are initialized to zero, the gradients are non-zero. The gradient calculation of the convolution can be formulated as follows:
\begin{equation}\small
	\left\{
	\begin{aligned}
        &\frac{\partial Z_c(\mathbf{C_f};\{\mathbf{W},\mathbf{B}\})}{\partial \mathbf{B}} = 1 \,, \\
        &\frac{\partial Z_c(\mathbf{C_f};\{\mathbf{W},\mathbf{B}\})}{\partial \mathbf{W}} = \mathbf{C_f} \,, \\ 
	\end{aligned}
	\right.
\end{equation} 
where $\mathbf{C_f}$ is the feature extracted from the image, ensuring it is non-zero. The gradient calculation of the linear layer is similar to the convolution layer. As the training progresses, the weights and biases of the convolution and linear layers gradually adjust to transform the detection features into an adaptive form for the recognition head.

\subsection{Adapter}
To enhance the synergy between the detector and recognizer in achieving joint optimization, we adopt the Adapter to fine-tune the model, inspired by ~\cite{houlsby2019parameter}. The architecture of the Adapter is illustrated in Fig.~\ref{fig:adapter}. We keep the parameters of the neural network locked (frozen), with the exception of the normalization layer. The purpose of the normalization layer is to adjust the mean and variance of the distribution of joint optimization features for the adapter. The adapter itself comprises two linear layers and an activation function~\cite{hendrycks2016gaussian}, as represented by the following equation:
\begin{equation}\small
\mathbf{f_o} = \varphi( \mathbf{f_i}\mathbf{W_1}^T+ \mathbf{B_1}) \mathbf{W_2}^T+ \mathbf{B_2} + \mathbf{f_i} \,,
\end{equation}
where $\mathbf{W_1}$, $\mathbf{W_2}$, $\mathbf{B_1}$ and $\mathbf{B_2}$ represent the parameters of the linear layers. $\varphi$ is the activation function. $\mathbf{f_i}$ and $\mathbf{f_o}$ represent the input and output. 

\begin{table*}[!ht]
\centering
\caption{\small Scene text spotting results on Total-Text and TextOCR. ``None'' refers to recognition without lexicon. ``Full'' lexicon contains all the words in the test set. DB+PARSeq, TESTR-det+MAERec, and DPText-DETR+DiG represent the two-step text spotting using the DB, TESTR's Detector, DPText-DETR as detector and PARSeq, MAERec, and DiG as the recognizer.}\label{tab:total_text}
\setlength{\tabcolsep}{10pt}
\resizebox{.9\linewidth}{!}{%
\begin{tabular}{lcl ccc cc c}
\toprule
\multirow{2}{*}{Method} & \multirow{2}{*}{Venue} & \multirow{2}{*}{Backbone} & \multicolumn{3}{c}{Detection} & \multicolumn{2}{c}{End-to-End} & \multirow{2}{*}{FPS}   \\
\cmidrule(lr){4-6} \cmidrule(lr){7-8}
& & & P & R & F & None & Full &  \\ 
\midrule
DB~\cite{liao2020real}+PARSeq~\cite{bautista2022parseq} & $-$ & ResNet-18 & 89.3 & 78.4 & 83.5 & 69.1 & 79.1 & 28.2 \\
TESTR-det~\cite{zhang2022text}+MAERec~\cite{jiang2023revisiting} & $-$ & ResNet-50 & 92.8 & 81.3 & 87.3 & 78.0 & 86.6 & 4.2 \\
DPText-DETR~\cite{ye2022dptext}+DiG~\cite{yang2022reading} & $-$ & ResNet-50 & 91.2 & 86.3 & 88.7 & 77.5 & 87.6 & 7.1 \\
Text Dragon \cite{feng2019textdragon} & ICCV'2019 & VGG16 & 85.6 & 75.7 & 80.3 & 48.8 & 74.8 & $-$ \\ 
Boundary TextSpotter \cite{wang2020all} & AAAI'2020 & ResNet-50-FPN & 88.9 & 85.0 & 87.0 & 65.0 & 76.1 & $-$ \\
Unconstrained \cite{qin2019towards} & ICCV'2019 & ResNet-50-MSF & 83.3 & 83.4 & 83.3 & 67.8 & $-$ & $-$ \\
Text Perceptron \cite{qiao2020text} & AAAI'2020 & ResNet-50-FPN  & 88.8 & 81.8 & 85.2 & 69.7 & 78.3 & $-$ \\
Mask TextSpotter v3 \cite{liao2020mask} & ECCV'2020 & ResNet-50-FPN & $-$ & $-$ & $-$ & 71.2 & 78.4 & $-$ \\ 
ABCNet \cite{liu2020abcnet} & CVPR'2020 & ResNet & $-$  & $-$ & 64.2 & 75.7 & 17.9 \\ 
ABCNet v2 \cite{liu2021abcnetv2} & TPAMI'2022 & ResNet-50-BiFPN & 90.2 & 84.1  & 87.0 & 70.4 & 78.1 & 10 \\ 
MANGO \cite{qiao2021mango} & AAAI'2021 &  ResNet-50-FPN & $-$ & $-$ & $-$ & 72.9 & 83.6 & 4.3 \\
PGNet \cite{wang2021pgnet} & AAAI'2021 & ResNet-50-FPN & 85.5 & 86.8   & 86.1 & 63.1 & $-$ & 35.5 \\
TESTR \cite{zhang2022text} & CVPR'2022 &ResNet-50 &93.4 &81.4 &86.9 &73.3 &83.9 &5.3 \\
TTS (poly) \cite{kittenplon2022towards} & CVPR'2022 & ResNet-50 &$-$ &$-$ &$-$ &78.2 &86.3 &$-$ \\
SwinTextSpotter \cite{huang2022swintextspotter} & CVPR'2022 &Swin-Tiny &$-$ &$-$ &88.0 &74.3 &84.1 &$-$ \\
ABINet++ \cite{fangabinet++} & TPAMI'2022 & ResNet-50-BiFPN & $-$ & $-$  & - & 77.6 & 
84.5 & 10.6 \\ 
SRSTS \cite{wu2022decoupling} & ACMMM'2022 & ResNet-50 &92.0 &83.0 &87.2 &78.8 &86.3 & 18.7\\
GLASS \cite{ronen2022glass} & ECCV'2022 & ResNet-50 &90.8 &85.5 & 88.1 & 79.9 &86.2 & 3.0 \\
SPTS v2 \cite{liu2023spts} & TPAMI'2023 & ResNet-50 &$-$ &$-$ &$-$ &75.5 &84.0 &$-$ \\
DeepSolo \cite{ye2023deepsolo} & CVPR'2023 & ResNet-50 &93.1 & 82.1 & 87.3 & 79.7 & 87.0 & 17.0 \\
UNITS \cite{kil2023towards} & CVPR'2023 & Swin-Base &$-$ & $-$ & 89.8 & 78.7 & 86.0 & $-$ \\
ESTextSpotter \cite{huang2023estextspotter} & ICCV'2023 & ResNet-50 & 92.0 &88.1 & 90.0 & 80.8 & 87.1 & 4.3 \\
\midrule
DG-Bridge Spotter & $-$ & ResNet-50 & 92.0 & 86.5 & 89.2 & \textbf{83.3} & \textbf{88.3} & 6.7 \\
\bottomrule
\label{tab:tt}
\end{tabular}
}
\end{table*}

\subsection{Optimization}
For optimization, we follow the original loss in the detector and recognizer. 
The loss function is the sum of detection loss $\mathcal{L}_{det}$ and recognition loss $\mathcal{L}_{rec}$, formulated as follows:
\begin{equation}\small
\mathcal{L}_{sum} = \lambda_{det}\mathcal{L}_{det} + \lambda_{rec}\mathcal{L}_{rec} \,,
\end{equation}
where $\lambda_{det}$ and $\lambda_{rec}$ represents a trade-off hyper-parameters and set to 1 in this paper. 

\section{Experiments}

\subsection{Implementation Details}
\label{Implementation Details}
We use the DPText-DETR~\cite{ye2022dptext} as the well-trained detector and the DiG~\cite{yang2022reading} as the well-trained recognizer in this paper, which is termed DG-Bridge Spotter. Official open-source weights for both the detector and recognizer are utilized. The adapter is incorporated into the Transformer encoder layer~\cite{zhu2020deformable} in the detector and decoder in the recognizer. We also attempt other combinations of detectors and recognizers to verify the effectiveness of our method, as described in Sec.~\ref{sec:ablation_studies}.
We utilize the AdamW~\cite{loshchilov2017decoupled} optimizer to optimize the Bridge and Adapter in DG-Bridge Spotter. The Bridge and Adapter are tuned on the training data of the target set. We train the Bridge and Adapter for 10,000 iterations, employing a batch size of eight images. The inference speed is tested on a single NVIDIA GeForce RTX 3090.
The data augmentation strategies are similar to those in prior works~\cite{zhang2022text,liu2020abcnet,liu2021abcnetv2}, as detailed below: 1) Random resizing is conducted with the shorter dimension ranging from $480$ to $832$ pixels, at intervals of $32$, while the longer dimension is constrained
within $1600$ pixels. 2) Random cropping is used, ensuring that text is not cut off. For testing, the shorter dimension of the image is resized to $1000$ pixels, while the longer dimension is constrained to a maximum of $1824$ pixels.

\subsection{Comparison with State-of-the-art Methods}
We evaluate our method on several benchmarks, including the multi-oriented benchmark ICDAR2015~\cite{karatzas2015icdar}, the word-level annotated arbitrarily-shaped text benchmark Total-Text~\cite{ch2020total}, and the line-level annotated arbitrarily-shaped text benchmark CTW1500~\cite{liu2019curved}. For various benchmarks, we replace only the detector and fine-tune the Bridge with the Adapter. In contrast, end-to-end text spotters require training the entire system. Unless otherwise stated, all values in the table are presented as percentages. We also present the results on the  TextOCR~\cite{singh2021textocr} and HierText~\cite{long2022towards}, in the supplementary material.

\begin{table}[!t]
    \centering
    \caption{\small End-to-end text spotting results on CTW1500. ``None'' represents lexicon-free, while ``Full'' indicates all the words in the test set are used.}\label{tab:ctw1500}
\resizebox{.9\linewidth}{!}{%
    \begin{tabular}{l ccc cc}
    \toprule
    \multirow{2}{*}{Method} & \multicolumn{3}{c}{Detection} & \multicolumn{2}{c}{End-to-End} \\ \cmidrule(lr){2-4} \cmidrule(lr){5-6} 
     & P & R & F & None & Full  \\ 
     \midrule
    Text Dragon \cite{feng2019textdragon} & 84.5 & 82.8 & 83.6 & 39.7 & 72.4 \\ 
    Text Perceptron \cite{qiao2020text} & 87.5 & 81.9 & 84.6 & 57.0 & $-$  \\
    ABCNet \cite{liu2020abcnet} & $-$ & $-$ & $-$ & 45.2 & 74.1  \\ 
    ABCNet v2 \cite{liu2021abcnetv2} & 85.6 & 83.8 & 84.7 &  57.5 & 77.2 \\
    MANGO \cite{qiao2021mango} & $-$ & $-$ & $-$ & 58.9 & 78.7  \\
    ABINet++ \cite{fangabinet++} & $-$ & $-$ & $-$ & 60.2 & 80.3  \\
    TESTR \cite{zhang2022text} & 92.0 & 82.6 & 87.1 & 56.0 & 81.5 \\
    SwinTextSpotter \cite{huang2022swintextspotter} & $-$ & $-$ & 88.0 & 51.8 & 77.0 \\
    SPTS v2 \cite{liu2023spts} & $-$ & $-$ & $-$ & 63.6 & \textbf{84.3} \\
    DeepSolo \cite{ye2023deepsolo} & $-$ & $-$ & $-$ & 64.2 & 81.4 \\
    ESTextSpotter \cite{huang2023estextspotter} & 91.5 & 88.6 & 90.0 & 64.9 & 83.9 \\
    \hline
    DG-Brigde Spotter  & 92.1	&86.2	&89.0	&\textbf{69.8}	& 83.9 \\
    \bottomrule
    \label{tab:ctw1500}
    \end{tabular}
}
\end{table}

\begin{figure}[!t]
    \centering
    \includegraphics[width=\linewidth]{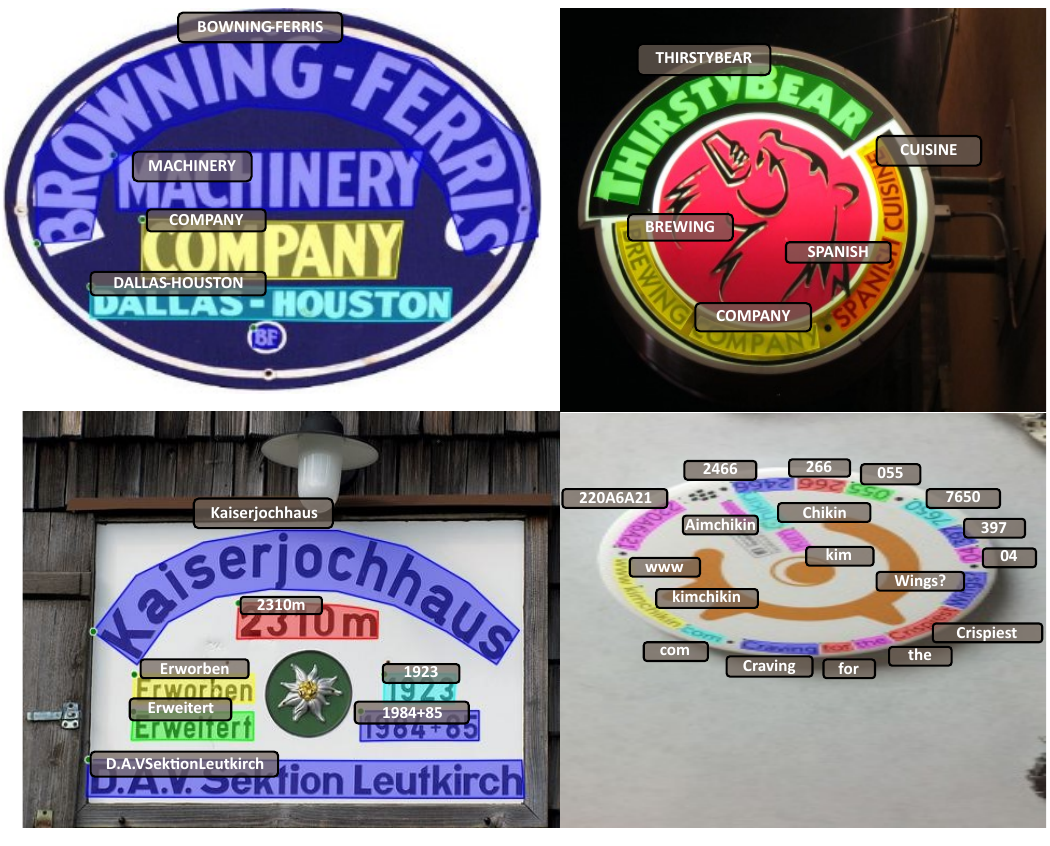}
    \caption{
    Qualitative results of DG-bridge Spotter on CTW1500 (left column) and Total-Text (right column). Zoom in for best view.
    }
    \label{fig:vis}
\end{figure}

\textbf{Total-Text.} For the word-level annotated arbitrarily-shaped text benchmark Total-Text, the results are presented in Tab.~\ref{tab:tt}. We find that our method achieves stronger improvement on the `None' Lexicon than `Full'. On the `None' lexicon, our method correctly recognizes the results that are incorrect in the baseline, leading to higher performance. However, on the `Full' lexicon, many incorrect results in the baseline can be corrected by the lexicon, resulting in comparable performance with ours. Therefore, our method shows better performance on the `None' lexicon compared to that on the `Full' lexicon. Some qualitative results are shown in Fig.~\ref{fig:vis}, demonstrating that our method is capable of recognizing text even in highly curved scenarios.

\textbf{CTW1500.} For the line-level annotated arbitrarily-shaped text benchmark CTW1500, the results are shown in Tab.~\ref{tab:ctw1500}. Our method surpasses the state-of-the-art method by a notable margin of $4.9\%$ in the `None' metrics, unequivocally proving its efficacy in handling long text recognition. Our method outperforms SPTS v2 and ESTextSpotter in `None' metrics while yielding comparable results in `Full' metrics. On the `None' lexicon, our method correctly recognizes the results that are incorrect in the baseline, leading to higher performance. This discrepancy highlights a prevalent issue in these methods, where a small number of characters within a text line are frequently misidentified, necessitating the use of a lexicon for correction. However, in many practical scenarios, lexicons are usually absent. 

\textbf{ICDAR2015.} Since DPText-DETR~\cite{ye2022dptext} does not provide open-source weights for the ICDAR2015 dataset, we adopt a similar method the detector of TESTR~\cite{zhang2022text} as our detector. It's important to highlight that we have kept the recognizer DiG~\cite{yang2022reading} unchanged. We refer to this combination of the detector of TESTR and the DiG-based recognizer as the TG-Bridge Spotter. The results are illustrated in Tab.~\ref{tab:icdar2015}. The proposed TG-Bridge Spotter outperforms the state-of-the-art method in all lexicons. Specifically, our method outperforms the ESTextSpotter~\cite{huang2023estextspotter}  by $1.6\%$, $1.2\%$ and $2.3\%$ in terms of `Strong', `Weak', and `Generic' metrics, respectively.

\begin{table}[!t]
\caption{\small Results on ICDAR 2015 dataset. ``S'', ``W'', ``G'' represent recognition with ``Strong'', ``Weak'' or ``Generic'' lexicon, respectively.}\label{tab:icdar2015}
\centering
\setlength{\tabcolsep}{3pt}
\resizebox{\linewidth}{!}{%
\begin{tabular}{l ccc ccc}
\toprule
\multirow{2}{*}{Method} & \multicolumn{3}{c}{Detection} & \multicolumn{3}{c}{End-to-End} \\
\cmidrule(lr){2-4} \cmidrule(lr){5-7}
 & P & R & F & S & W & G  \\
\midrule
FOTS \cite{liu2018fots} & 91.0 & 85.2 & 88.0 & 81.1 & 75.9 & 60.8  \\
CharNet R-50 \cite{xing2019convolutional} & 91.2 & 88.3 & 89.7 & 80.1 & 74.5 & 62.2  \\
Boundary TextSpotter \cite{wang2020all} & 89.8 & 87.5 & 88.6 & 79.7 & 75.2 & 64.1  \\
Unconstrained \cite{qin2019towards} & 89.4 & 85.8 & 87.5 & 83.4 & 79.9 & 68.0  \\
Text Perceptron \cite{qiao2020text} & 92.3 & 82.5 & 87.1 & 80.5 & 76.6 & 65.1  \\
Mask TextSpotter v3 \cite{liao2020mask} & $-$ & $-$ & $-$ & 83.3 & 78.1 & 74.2 \\
ABCNet v2 \cite{liu2021abcnetv2} &  90.4 & 86.0 & 88.1 & 82.7 & 78.5 & 73.0  \\
MANGO \cite{qiao2021mango} & $-$ & $-$ & $-$ & 81.8 & 78.9 & 67.3  \\
PGNet \cite{wang2021pgnet} & 91.8 & 84.8 & 88.2 & 83.3 & 78.3 & 63.5 \\
ABINet++ \cite{fangabinet++} &$-$ &$-$ &$-$ &84.1 & 80.4 & 75.4 \\
TESTR \cite{zhang2022text} &90.3 &89.7 &90.0 &85.2 &79.4 &73.6 \\
TTS \cite{kittenplon2022towards}&$-$ &$-$ &$-$ &85.2 &81.7 & 77.4 \\
SwinTextSpotter \cite{huang2022swintextspotter} &$-$ &$-$ &$-$ &83.9 &77.3 &70.5 \\
SPTS v2 \cite{liu2023spts} &$-$ &$-$ &$-$ &82.3 &77.7 &72.6\\
SRSTS \cite{wu2022decoupling} &96.1 &82.0 &88.4 &85.6 &81.7 &74.5  \\
GLASS \cite{ronen2022glass}  &86.9 &84.5 &85.7 &84.7 &80.1 &76.3 \\
DeepSolo \cite{ye2023deepsolo} &92.8 &87.4 & 90.0 & 86.8 &81.9 &76.9  \\
ESTextSpotter \cite{huang2023estextspotter}  & 92.5 & 89.6 &  91.0 & 87.5 & 83.0 & 78.1 \\
\midrule
TG-Bridge Spotter & 93.8 & 87.5 & 90.5 & \textbf{89.1} & \textbf{84.2} & \textbf{80.4}  \\
\bottomrule
\end{tabular}
}
\end{table}

\subsection{Ablation Studies}
\label{sec:ablation_studies}
We conduct ablation studies on Total-Text with three combinations of detectors and recognizers to verify the validity of our method. (1) we adopt the DPText-DETR~\cite{ye2022dptext} and the DiG~\cite{yang2022reading}, termed DG-Bridge Spotter. (2) we attempt another combination that involves the detector of TESTR~\cite{zhang2022text} and MAERec~\cite{jiang2023revisiting}, termed TM-Bridge Spotter. (3) we also utilize the combination of segmentation-based text detector DBNet~\cite{liao2020real} as the detector and a fast decoding recognizer PARSeq~\cite{bautista2022parseq}, dubbed BP-Bridge Spotter. We choose the ResNet18~\cite{he2016deep} as the backbone of the DBNet.

\begin{table}[!t]
\centering
\caption{\small Ablation study on Total-Text. DA means using the adapter in the detector. RA means using the adapter in the recognizer. DG/TM/BP-Baseline represent the corresponding two-step pipelines.}\label{tab:ablation_modules}
\setlength{\tabcolsep}{4pt}
\resizebox{\linewidth}{!}{%
\begin{tabular}{cc ccccc c cc}
\toprule
 \multirow{2}{*}{Method} & \multirow{2}{*}{Bridge} & \multirow{2}{*}{DA} & \multirow{2}{*}{RA} & \multicolumn{3}{c}{Detection} & \multirow{2}{*}{E2E} & \multirow{2}{*}{FPS} & \multirow{2}{*}{Param} \\
  \cmidrule(lr){5-7}
 &  & &  & P & R & F &  \\
 \midrule
 DG-Baseline & $-$ & $-$ & $-$ & 91.2 & 86.3 & 88.7 & 77.5 & 7.1 & 81.5M  \\
 DG-Baseline+ & \checkmark & $-$ & $-$ & 91.2 & 86.3 & 88.7 & 81.5 & 7.1 & 85.0M \\
 DG-Baseline+ & \checkmark & \checkmark & $-$ & 91.7 & 86.7 & 89.1 & 82.1 & 7.0 & 85.2M \\
 DG-Bridge Spotter & \checkmark & \checkmark & \checkmark & 92.0 & 86.5 & 89.2 & \textbf{83.3} & 6.7 & 86.0M \\
 \midrule
 TM-Baseline & $-$ & $-$ & $-$ & 92.8 & 81.3 & 87.3 & 78.0 & 4.2 & 77.4M \\
 TM-Baseline+ & \checkmark & $-$ & $-$ & 92.8 & 81.3 & 87.3 & 80.8 & 4.2 & 80.9M \\
 TM-Baseline+ & \checkmark & \checkmark & $-$ & 92.3 & 83.6 & 87.8 & 81.1 & 4.0 & 81.1M \\
 TM-Bridge Spotter & \checkmark & \checkmark & \checkmark & 92.4 & 82.7 & 87.3 & \textbf{81.9} & 3.6 & 81.5M\\
 \midrule
 BP-Baseline & $-$ & $-$ & $-$ & 89.3 & 78.4 & 83.5 & 69.1 & 28.2 & 37.6M\\
 BP-Baseline+ & \checkmark & $-$ & $-$ & 89.3 & 78.4 & 83.5 & 70.7 & 28.0 & 40.4M \\
 BP-Baseline+ & \checkmark & \checkmark & $-$ & 88.8 & 79.7 & 84.0 & 70.8 & 27.0 & 40.5M \\
 BP-Bridge Spotter & \checkmark & \checkmark & \checkmark & 89.1 & 79.1 & 83.8 & \textbf{72.5} & 26.5 & 40.9M\\
 \bottomrule
 \label{tab:bridge}
\end{tabular}%
}
\end{table}

\begin{figure}[t]
    \centering
    \includegraphics[width=\linewidth]{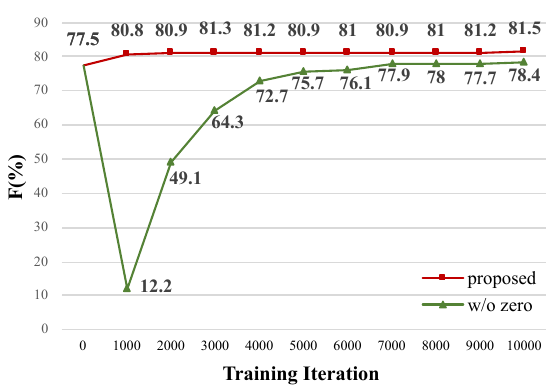}
    \caption{
    Ablative study of the zero-initialized weight in Bridge. “F” indicates F-measure in end-to-end text spotting results on Total-Text.
    }
    \label{fig:ablation}
\end{figure}

\begin{figure}[!t]
    \centering
    \includegraphics[width=\linewidth]{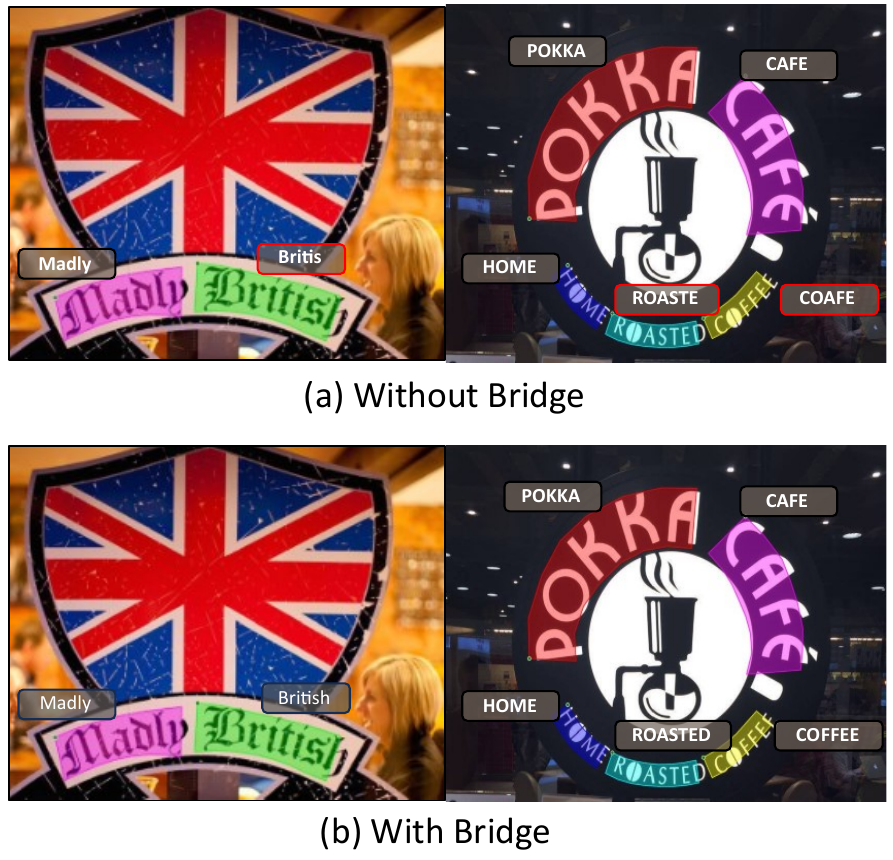}
    \caption{
    Effectiveness of Bridge. Red boxes indicate recognition errors due to inaccurate detection results. Zoom in for best view.
    }
    \label{fig:erro_accumulation}
\end{figure}

\paragraph{Ablation Study of The Bridge.}
To evaluate the effectiveness of the proposed Bridge and Adapter, we conduct ablation studies on the Total-Text. As shown in Tab.~\ref{tab:bridge},  Bridge significantly improves text spotting performance in all three combinations, with a $4.0\%$ increase in DG-Bridge Spotter, a $2.8\%$ increase in TM-Bridge Spotter, and $1.6\%$ increase in BP-Bridge Spotter. The Bridge successfully merges detection features with large receptive fields and recognition features with high resolution, directing this combined information into a locked recognition head. This process enhances performance significantly, with only a slight reduction in speed and requiring minimal increases in parameters. Moreover, we offer a more intuitive comparison to highlight the efficacy of the Bridge, as depicted in Fig.~\ref{fig:erro_accumulation}. In scenarios without the Bridge, if the detection results do not align accurately with the text boundaries, it can readily result in inaccurate recognition results. This situation is commonly known as error accumulation. In contrast, the Bridge effectively alleviates this issue by utilizing features with large receptive fields, thereby improving the performance.

\begin{table}[!t]
\centering
\caption{\small Ablation study about different Transformer layers in the Bridge.}\label{tab:ablation_modules}
\setlength{\tabcolsep}{4pt}
\resizebox{0.8\linewidth}{!}{%
\begin{tabular}{cc cc c c c }
\toprule
 \multirow{2}{*}{Num layer}  & \multicolumn{3}{c}{Detection} & \multirow{2}{*}{E2E} & \multirow{2}{*}{FPS} & \multirow{2}{*}{Param}\\
  \cmidrule(lr){2-4}
 & P & R & F &  \\
 \midrule 
 0  & 91.3 & 86.0 & 88.6 & 82.3 & 2.9 & 84.2M \\
 1 & \textbf{92.0} & 86.5 & \textbf{89.2} & \textbf{83.3} & 2.9 & 86.0M \\
 3  & 91.1 & \textbf{86.6} & 88.8 & 82.6 & 2.9 & 89.6M \\
 6 & 91.8 & 86.5 & 89.1 & 82.9 & 2.9 & 94.9M \\
 \bottomrule
 \label{tab:tr_layers}
\end{tabular}%
}
\end{table}

\paragraph{Ablation Study of The Adapter.}
The influence of Adapter is depicted in Tab.~\ref{tab:bridge}. Through the integration of the Adapter into both the detector and recognizer, three combinations effectively acquire end-to-end optimization features, thereby improving overall performance. Furthermore, it introduces only a modest number of additional parameters and has a minimal impact on speed.

\paragraph{Ablation Study of the Zero-initialized Weight in Bridge.}
To further investigate the efficacy of the proposed zero-initialized architecture, we conduct an analysis of the alternative Bridge structure by substituting the zero-initialized convolution and linear layers with counterparts initialized using Gaussian weights. In order to control variables, we do not add Adapter in this experiment. The results are illustrated in Fig.~\ref{fig:ablation}. The findings reveal that the zero-initialized weight within the Bridge facilitates rapid adaptation of the recognition head to features with large receptive fields, leading to a substantial performance boost (from $77.5\%$ to $81.5\%$). In contrast, employing a Gaussian-initialized structure initially disrupts the training of the recognition head, resulting in only marginal eventual improvement (from $77.5\%$ to $78.4\%$).

\paragraph{Ablation Study of The Number of Transformer Layers.}
To comprehensively validate the impact of the number of Transformer layers in the Bridge, we conduct experiments on Total-Text to compare the performance with varying numbers of Transformer layers in the Bridge. As presented in Tab.~\ref{tab:tr_layers}, we observe that setting the number to 1 resulted in the model achieving the best performance. Despite reducing the number of parameters by $1.8 M$, the detection performance dropped by $0.6\%$, and the text spotting performance decreased by $1\%$. Further increasing the number of Transformer layers does not lead to performance improvement, but it does increase the model's parameters.

\begin{table}[!tb]
\centering
\caption{\small Comparison of different paradigms on Total-Text. Two-step finetune represents fine-tuning the detector and recognizer, respectively, without Bridge. Row3+Using $C_f$ represents using the $C_f$ as the input of the recognizer. Row4+Using $C_i$ represents using $C_f$ and $C_i$ as the input of the recognizer.
}
\label{tab:ablation_modules}
\setlength{\tabcolsep}{4pt}
\resizebox{\linewidth}{!}{%
\begin{tabular}{cc cc c c c }
\toprule
 \multirow{2}{*}{Method}  & \multicolumn{3}{c}{Detection} & \multirow{2}{*}{E2E} & \multirow{2}{*}{FPS} & \multirow{2}{*}{Param}\\
  \cmidrule(lr){2-4}
 & P & R & F &  \\
 \midrule
 Two-step & 91.2 & 86.3 & 88.7 & 77.5 & 7.1 & 81.5M \\
 End-to-end  & 91.2 & 86.1 & 88.6 & 75.6 & 7.5 & 83.1M \\
 Two-step finetune  & 89.4 & 85.7 & 87.5 & 78.8 & 7.1 & 81.5M \\
 Row3+Using $C_f$  & 91.2 & 86.5 & 88.8 & 66.8 & 7.5 & 83.1M\\
 Row4+Using $C_i$  & 91.0 & 87.7 & 89.3 & 79.8 & 7.1 & 83.1M\\
 Ours & 92.0 & 86.5 & 89.2 & \textbf{83.3} & 6.7 & 85.2M \\
 \bottomrule
 \label{tab:different_combinations}
 \end{tabular}%
 }
\end{table}

\paragraph{Comparison with Existing Paradigm.}
To further validate the effectiveness of our method, we conduct experiments on Total-Text, comparing it with two-step and end-to-end text spotting. This experiment uses the DPText-DETR as the detector and DiG as the recognizer. The results are illustrated in Tab.~\ref{tab:different_combinations}. In the case of end-to-end text spotting, we made a modification by adjusting the first layer of the recognition backbone. This adjustment enables a seamless passage of cropped features from the detection backbone to the recognizer. We observed that when using the same detector and recognizer, two-step one outperforms end-to-end one in text spotting metrics, despite their comparable detection performance. This superiority can be attributed to the high modularity of the two-step method. This modularity enables the recognizer to be trained independently with a larger dataset, potentially surpassing the impact of error accumulation and sub-optimization. Additionally, we made an effort to load pre-trained weights in a two-step method to initialize both the detector and recognizer. We also make efforts to fine-tune the detector and recognizer directly on TotalText and connect them in a two-step manner, as illustrated in the third row of Tab.~\ref{tab:different_combinations}. Due to its limitations in effectively mitigating the issues of sub-optimal performance and error accumulation, this approach offers only marginal performance improvements. Subsequently, we load the pre-trained weight of the detector and recognizer and use the $C_f$ as the input of the recognizer, referred to as Row3+Using $C_f$. The $C_f$ and $C_i$ are detailed in the Sec.~\ref{sec:overall}. As shown in Tab.~\ref{tab:different_combinations}, the results indicate a drop in performance. This is because, during pre-training, the inputs of the recognizer in Row3+Using $C_f$ are the images. Consequently, when the input transitions to features in detection, the recognizer misinterprets them as noise, disrupting the fine-tuning process of the recognizer. We further utilize the summation of $C_f$ and $C_i$ as the input of the recognizer, referred to as Row4+Using $C_i$. The results demonstrate the effectiveness of integrating both detection and recognition features. 

\section{Conclusion}

In this paper, we introduce a new paradigm for text spotting, termed Bridging Text Spotting, to address the issues of sub-optimal performance and error accumulation in the two-step text spotting while retaining modularity. The proposed Bridge effectively connects the well-trained detector and recognizer through a zero-initialized neural network. The Adapter enables the well-trained detector and recognizer to efficiently learn end-to-end optimization features, thereby improving performance. Extensive experimental results demonstrate the effectiveness of Bridging Text Spotting: 1) Bridging Text Spotting with the latest detector and recognizer outperforms the previous state-of-the-art end-to-end method on various challenge benchmarks. 2) Bridging Text Spotting can consistently enhance performance across various combinations of detectors and recognizers. The proposed method provides an effective way of integrating two distinct modules for end-to-end optimization. In the future, how to connect multi-task modules using our approach to create a robust multi-task system is worthy of further study. 

\noindent \paragraph{Acknowledgement} This research is supported in part by National Natural Science Foundation of China (Grant No.: 62441604, 61936003), National Key Research and Development Program of China  (2022YFC3301703)

\maketitle
\thispagestyle{empty}
\appendix
{\centering\section*{Appendix}}

\section{Performance on TextOCR and HierText}

We conduct experiments on more challenging benchmarks on TextOCR~\cite{singh2021textocr} and HierText~\cite{long2022towards} to verify the effectiveness of our methods. For TextOCR, we adopt a detector similar to the GLASS~\cite{ronen2022glass} and use the DiG~\cite{yang2022reading} as the recognizer. For HierText, we utilize the DBNet++~\cite{liao2022real} as the detector and MAERec~\cite{jiang2023revisiting} as the recognizer. As shown in Tab.~\ref{tab:textocr_hiertext}, the results demonstrate the effectiveness of our method.

\begin{table}[!h]
    \centering
    \caption{\small End-to-end text spotting results on TextOCR and HierText.}\label{tab:ctw1500}
\resizebox{.9\linewidth}{!}{%
    \begin{tabular}{l cc}
    \toprule
    \multirow{2}{*}{Method} & \multirow{2}{*}{TextOCR} & \multirow{2}{*}{HierText} \\
     &  &     \\ 
     \midrule
    MaskTextSpotter v3 \cite{liao2020mask} & 50.8 & $-$ \\
    GLASS \cite{ronen2022glass} & 67.1 & $-$ \\
    HTS \cite{long2024hierarchical} & $-$ & 75.6 \\
    \hline
    Ours  & 68.5 &76.1 \\
    \bottomrule
    \label{tab:textocr_hiertext}
    \end{tabular}
}
\end{table}

{
    \small
    \bibliographystyle{ieeenat_fullname}
    \bibliography{main}
}
\end{document}